\documentclass[11pt,a4paper]{article}
\usepackage[hyperindex,breaklinks]{hyperref}
\usepackage[hyperref]{emnlp2020}
\usepackage{times}
\usepackage{latexsym}
\usepackage{times}
\usepackage{url}
\usepackage{latexsym}
\usepackage{array}
\usepackage{pgfplots}
\usepackage{graphicx} 
\usepackage{subcaption}
\usepackage{makecell}
\usepackage{xcolor}
\usepackage{blindtext}%

\usepackage{microtype}

\aclfinalcopy 

\definecolor{darkgreen}{RGB}{17,99,17}

\title{Evaluating Transformer-Based Multilingual Text Classification}

\author{Sophie Groenwold\thanks{\hspace{1 mm} Equal contribution.}, Samhita Honnavalli\footnotemark[1], Lily Ou\footnotemark[1], Aesha Parekh\footnotemark[1], \\ \textbf{Sharon Levy, Diba Mirza, William Yang Wang} \\
University of California, Santa Barbara \\
\texttt{ \{sophiegroenwold, shonnavalli, lilyou, aeshaparekh\}@ucsb.edu }
\\
\texttt{ \{sharonlevy, dimirza, william\}@cs.ucsb.edu }
}

\date{}

\begin{document}
\maketitle
\begin{abstract}

As NLP tools become ubiquitous in today's technological landscape, they are increasingly applied to languages with a variety of typological structures. However, NLP research does not focus primarily on typological differences in its analysis of state-of-the-art language models. As a result, NLP tools perform unequally across languages with different syntactic and morphological structures. Through a detailed discussion of word order typology, morphological typology, and comparative linguistics, we identify which variables most affect language modeling efficacy; in addition, we calculate word order and morphological similarity indices to aid our empirical study. We then use this background to support our analysis of an experiment we conduct using multi-class text classification on eight languages and eight models. 



\end{abstract}

\section{Introduction}

Historically, NLP research has focused on a select group of English-similar languages for which there exists abundant training data, and thus favors typological characteristics shared by those languages \cite{joshi2020state}. However, the increasing prevalence of NLP technologies has made it imperative that models perform equally across all languages, regardless of a language's text corpora availability. 

Current NLP tools are multilingual in the sense that there is nothing explicitly preventing them from being utilized for different languages given that there exist annotated data \cite{mielke-etal-2019-kind}. However, this does not mean that NLP tools perform fairly across all languages. While it is generally recognized that linguistic properties should be taken into account during the creation of NLP models \cite{Bender:2009:LNL:1642038.1642044}, there is little consensus on which linguistic features most influence a language's modeling performance. Therefore, we seek to identify these typological attributes. 

Towards this end, we first provide a comprehensive background of linguistic aspects that we hypothesize affect language modeling efficacy, broken down into three main categories: word order typology, morphological typology, and comparative linguistics (the comparison of languages based on language family). To quantify our observations from the first two categories, we borrow a metric from the related linguistics literature: similarity index, which is used to describe how similar a target language's features are to English features. We calculate similarity indices for both word order and morphological features. 

We then apply these observations to the results of an empirical study we carry out that demonstrates the performance of eight models on the task of text classification, each trained on texts in eight languages (Chinese, English, French, German, Italian, Japanese, Russian, and Spanish) of the MLDoc corpus \cite{SCHWENK18.658}.

Of the metrics we consider, we find that language family provides the best indication of language modeling results. We conclude that it remains imperative for NLP researchers to consider the typological properties outlined, while measures of linguistic similarity between two languages must be refined.

Thus, our contributions include:
\begin{itemize}
  \item[$\bullet$] A summary of significant linguistic features, written from a perspective pertinent to NLP, and a categorization of major languages based on those features.
  \item[$\bullet$] A comprehensive experiment that demonstrates a disparity in the performance of eight models, where each model is trained and tested on a single language for varying sizes of data with equal label distribution.

  \item[$\bullet$] An analysis of language modeling efficacy in the contexts of word order typology, morphological typology, and comparative linguistics, including an illustration of the relationship between language-pair similarity and performance.
\end{itemize}

\section{Related Work}

\newcite{mielke-etal-2019-kind} investigates NLP language-agnosticism by evaluating a difficulty parameter with recurrent neural network language models. This work has found that word inventory size, or the number of unique tokens in the training set, had the highest correlation with their difficulty parameter \cite{mielke-etal-2019-kind}. This is in opposition to their previous hypothesis that morphological counting complexity (MCC), a simple metric to measure the extent to which a given language utilizes morphological inflection, is a primary factor in determining language modeling performance \cite{cotterell-etal-2018-languages}. We wish to build on this work by introducing increasingly nuanced indices for measuring language-pair similarity in both word order and morphological aspects.

Much of the prior research in analyzing typology from a modeling perspective focuses on the linguistic learning of BERT \cite{devlin-etal-2019-bert}, and as much of our analysis is also centered on our BERT results, we will do so as well here. Previous work has sought to pinpoint what typological aspects BERT learns through cross-lingual transfer tasks. \newcite{pires2019multilingual} evaluates mBERT in this context and hypothesizes that lexical overlap and typological similarity improve cross-lingual transfer. To assess typological similarity, they use six word order features adopted from WALS \cite{pires2019multilingual, wals}. In this work, the authors note that mBERT transfers between languages with entirely different scripts, and thus no lexical overlap. Building on this, \newcite{k2019crosslingual} disproves that word-piece similarity, or lexical overlap, improves or otherwise affects cross-lingual abilities. Instead, they propose that structural similarities are the sole language attributes that determine modeling efficacy, but do not elaborate on the specific linguistic properties that may be (word order, morphological typology, and word frequency, for example) \cite{k2019crosslingual}.

In this paper, we will build upon the aforementioned work by evaluating a text classification task on eight languages, where for each task we use the same language for both training and testing. This allows us to pinpoint linguistic features that aid modeling in a general context, as opposed to a cross-lingual context. Additionally, we expand our discussion of linguistics to include both word order and morphological typology, for both of which we will quantify by using a metric from the related linguistic literature.

\section{Linguistic Background}

We approach our analysis of the language modeling disparity from a linguistic perspective; as such, we break our discussion into three main areas: word order typology, morphological typology, and comparative linguistics. For further research, it may help the reader to know that the former two lie within the subfield of linguistic typology (which examines languages based on their structural features), whereas the latter (which studies the historical development and family categorization of languages) is a different subfield of linguistics.

\begin{table*}
        \centering
            \begin{tabular}{ |c|c|c|c|c| }
                \hline
                \textit{Language} & 
                    \textit{Constituent Order} & 
                    \textit{\makecell{Analytic \\ or Synthetic}} & 
                    \textit{\makecell{Agglutinative \\ or Fusional}} & 
                    \textit{Language Family} \\ \hline
                English & SVO & Analytic & N/A & Indo-European: Germanic
                    \\ \hline 
                Chinese & SVO & Analytic & N/A & Sino-Tibetan: Sinitic
                    \\ \hline 
                French & SVO & Synthetic & Fusional & Indo-European: Romance
                    \\ \hline 
                Italian & SVO & Synthetic & Fusional & Indo-European: Romance
                    \\ \hline 
                Spanish & SVO & Synthetic & Fusional & Indo-European: Romance 
                    \\ \hline 
                Japanese & SOV & Synthetic & Agglutinative & Japonic: Japanese 
                    \\ \hline 
                Russian & SVO & Synthetic & Fusional & Indo-European: Slavic 
                    \\ \hline 
                German & SOV \& SVO & Synthetic & Fusional & Indo-European: Germanic 
                    \\ \hline
            \end{tabular}
        \caption{Major topics of discussion in Section 3 by language, ordered by word order from most rigid (English) to most flexible (German).}
\end{table*}

\subsection{Word Order}

Word order typology is the study of the ordering of words within a sentence. In our discussion of word order, we will focus chiefly on constituent word order, where a constituent is defined as a stand-alone unit of language (for example, a word or phrase); however, within word order typology there do exist other points of analysis, such as modifier order, that we will not consider due to their perceived limited effect on modeling efficacy. We will also discuss word order flexibility, which is the frequency at which a sentence will vary from its dominant order. A summary of these details can be found in Table 1. 

\subsubsection{Constituent Order}

Nearly all languages have a dominant word order that is the most frequently used ordering of a sentence's subject, verb, and object \cite{comrie1989language}. Given the flexibility of word order in a certain language (see Section 3.1.2 below), a sentence in that language might vary from the dominant order because of situational constraints, to place emphasis, or to convey emotion. Since the dataset we use in our empirical study is composed of formal news stories, we expect our samples to adhere to each language's dominant word order more than what is perhaps representative of the language in other contexts (e.g. informally).

The constituent orders present in this dataset are subject-verb-object (SVO) for English, Spanish, French, Russian, Italian, and Chinese and subject-object-verb (SOV) for Japanese. German is not categorized as either. There are other word ordering schemes present in the world's languages, such as verb-subject-object (VSO) and object-verb-subject (OVS) that are not represented in this corpus.

We note two language-specific details here. Although most languages can be classified into a single dominant word order, the degree of ``dominance” the dominant word order has is related to the language's word order flexibility. Although we will discuss word order flexibility shortly, it is important to note that Russian, as a language with flexible word order, is still considered an SVO language despite that ordering being less strict than other languages \cite{wals-81}. However, German varies between SVO (in main clauses without an auxiliary verb) and SOV (in clauses with an auxiliary verb and in subordinate clauses) equally and is thus categorized as lacking a dominant verb order \cite{wals-81}.

\subsubsection{Word Order Flexibility}

Without delving too deep into the nuances of word order flexibility, we wish to provide some general observations. For our purposes, we simplify word order flexibility to a spectrum with rigid word order (where the constituent and modifier word orders follow strict grammatical rules) on one end and flexible word order (where morphological marking is grammatically necessary) on the other \cite{wals-81,59036}. However, when discussing word order flexibility it is imperative to identify which specific elements are flexible, and thus we include this when necessary.

Of the languages in this dataset, English has the most rigid word order, as modern English always follows a subject-verb format, which is usually realized in an SVO setting \cite{Bozsahin}. Chinese follows similar constituent order patterns to English, with greater degrees of flexibility in verb-object order \cite{Qian}. Next, French, Italian and Spanish are generally regarded as having similar levels of word order flexibility; however, related work views the three languages in terms of their grammatization, and thus we can view French as the most rigid of the three, followed by Italian and then Spanish, though the margins of difference between the three are low \cite{Lahousse}. Japanese has more flexibility than the languages discussed previously, as it frequently uses both SOV and OSV word orders \cite{Bozsahin}. Finally, as mentioned above, Russian depends heavily on morphological marking despite following an SVO constituent order, and German does not follow a single constituent order at all, thus making it the most flexible \cite{wals-81}.

\subsection{Morphological Typology}

While in the previous section we used word order typology to look at the ordering of constituents within a sentence or phrase, we now turn to morphological typology to examine the internal structure of words, i.e. the patterns that involve the creation and structure of words, primarily by their morphemes \cite{genetti2014}. A morpheme is the smallest semantic unit in language, and differs from a constituent in that it cannot convey meaning alone; for instance, the word ``unbelievable” would be considered a constituent, whereas ``un,” ``believe,” and ``able” are morphemes. 

Having a thorough understanding of morphemes and the different ways they are utilized across languages is essential for researchers in NLP because many contemporary NLP models use word embeddings as the underlying representation for language, using primarily Word2Vec \cite{mikolov2013efficient} or GloVe \cite{pennington-etal-2014-glove} architectures. Recent work in this area has developed word embedding methods based on morphemes, instead of constituents \cite{luong-etal-2013-better,cotterell-etal-2016-morphological-smoothing}. Considering the prevalence of these methods and their importance in word embedding processes, we further investigate the morphological diversity of the languages in this dataset.

Traditionally, linguists have sought to classify languages morphologically along two axes: analytic/synthetic and agglutinative/fusional \cite{Garland_morphologicaltypology}. The former describes the number of morphemes per word, and the latter the clarity of distinctions between morphemes. A summary of these details can be found in Table 1.

\subsubsection{Analytic - Synthetic Axis}

Given that inflection is a modification of a word for a particular grammatical category, we can define an analytic language as containing fewer morphemes per word and using less inflection, and a synthetic language as containing more morphemes per word and using more inflection. Of this dataset, English and Chinese are analytic, while the remaining languages fall within the synthetic spectrum. Chinese is more so than English, as it formerly was seen as an isolating language (an extreme of analytic, where there is no use of inflection and words are little more than morphemes); however, today it is considered to be analytic because its words commonly contain more than one morpheme \cite{bybee_1997}. English is more moderately analytic and is thus commonly categorized as fusional, which is typically reserved for synthetic languages.

Because analytic languages use little inflection, information is instead conveyed through tools such as word order, “helper” words, and context. Thus, there is a correlation between increased word order rigidity and the analytic categorization.

\subsubsection{Agglutinative - Fusional Axis}

Within synthetic languages, agglutinative languages have easily discernible morphemes, and morphemes typically contain a single feature; fusional languages use a single inflectional morpheme to signify multiple grammatical meanings. Six languages in this dataset--French, Spanish, Italian, Russian, German, and Japanese--are synthetic, where all are fusional except Japanese which is agglutinative. Therefore, Japanese follows a more regular morphology, whereas the fusional languages may be considered more morphologically complex (with German, then Russian having the highest degrees of inflection).

\subsection{Comparative Linguistics}

Another beneficial angle to compare languages from is comparative linguistics, which traces the development of modern natural languages through historical comparison and categorizes them based on common origin into language families. 

First, we examine the families of the languages of this dataset. Of the eight languages we used, six belong to the West European branch of the Indo-European language family; English and German are Germanic languages; Spanish, Italian, and French are Romance languages; and Russian is a Slavic language. Chinese descends from the Sino-Tibetan family, and Japanese from the Japonic family \cite{Ethnologue}.

This dataset acutely under-represents languages from other language families. For instance, West European languages are but one branch of the Indo-European superfamily; a significant sister family to the West European languages is the Indo-Iranian branch (which includes Persian from the Iranian sub-family and Hindi from the Indic subfamily). We also have no languages from the Niger-Congo and Afro-Asiatic families (which are the two major superfamilies of Africa and together make up 27\% of the known world languages) the Austronesian families (at 17.7\%), the Trans-New Guinea families (at 6.8\%), or considerable others \cite{Ethnologue}. 

Looking at familial origins is beneficial for easy categorization of languages; however, they are not always indicative of typological properties, which are intuitively more important for a given language's modeling efficacy. For example, although both English and German are classified as Germanic languages, they are different when compared in both word order and morphological typology (see Table 1); this is because English has come into areal contact with more languages than German, which is considered more archaic. Thus, Modern English typology more closely resembles French typology, as it has more recent influence from French and other Romance languages than from its Germanic roots.

\section{Measures of Language Similarity}

For clarity of analysis, we use two similarity indices that we have adopted and modified from the related linguistic literature \cite{comrie_2016}. We use the relevant features from the World Atlas of Language Structures, or WALS, where a feature is defined as a structural property of language that differs across languages \cite{wals}. For each similarity index, we use only WALS entries that are categorized as ``morphology" or ``word order" features, as appropriate. 

\begin{table}
        \centering
            \begin{tabular}{ |c|c| }
                \hline
                \makecell[l]{English - Russian} & 92.86 \\ \hline
                \makecell[l]{English - Spanish} & 86.21 \\ \hline
                \makecell[l]{English - Italian} & 85.19 \\ \hline
                \makecell[l]{English - Chinese} & 70.73 \\ \hline
                \makecell[l]{English - French} & 65.52 \\ \hline 
                \makecell[l]{English - German} & 51.72 \\ \hline 
                \makecell[l]{English - Japanese} & 30.44 \\
                \hline
            \end{tabular}
        \caption{Word order similarity indices of the languages in the MLDoc corpus, from greatest to least.}
\end{table}

If there is any difference in language performance across NLP models, we expect it to correlate with how similar a language is to English. We anticipate this simply because many NLP technologies have been made with the underlying assumption of English typology. Thus, to measure English-similarity, evaluate this hypothesis, and identify the typological characteristics that have the greatest correlation with modeling performance, we utilize two similarity metrics: word order similarity index and morphological similarity index.

Each similarity index is determined using WALS data and is computed as follows. Given two languages, we first count the number of features that are documented for both; we then count the instances of equal categorization for each feature. For instance, in finding the English - Chinese morphological similarity index, we identify 12 morphology features that WALS has recorded for both, and six that contain the same value. Thus the English - Chinese morphological similarity index is 50. A complete table of word order similarity indices and morphological similarity indices between English and the remaining seven languages in this dataset can be found in Tables 2 and 3, respectively.\footnote{English is excluded, as the similarity indices displayed in the table are computed with respect to English.}

We note that it is perhaps problematic to measure languages based on their similarity to English. However, we do so to demonstrate an existing disparity in language modeling efficacy; by no means are we suggesting that it is ethical for English to be the accepted standard in broader applications of NLP.

\begin{table}
        \centering
            \begin{tabular}{ |c|c| }
                \hline
                \makecell[l]{English - German} & 83.33 \\ \hline 
                \makecell[l]{English - Russian} & 66.67 \\ \hline 
                \makecell[l]{English - French} & 58.33 \\ \hline 
                \makecell[l]{English - Japanese} & 58.33 \\ \hline
                \makecell[l]{English - Spanish} & 50 \\ \hline 
                \makecell[l]{English - Chinese} & 50 \\ \hline 
            \end{tabular}
        \caption{Morphological similarity indices of the languages in the MLDoc corpus, from greatest to least.\protect\footnotemark}
\end{table}

\footnotetext{We do not include a morphological similarity index for Italian, as there are only two WALS morphology features encoded for both Italian and English.}

\begin{figure*}
    \begin{tikzpicture}
        \begin{axis}[
                width=0.95\textwidth,
                height=0.4\textwidth,
                ybar,
                ymin=0.88,
                ymax = 0.98,
                ytick={0.88, 0.9, 0.92, 0.94, 0.96, 0.98},
                yticklabels={0.88, 0.9, 0.92, 0.94, 0.96, 0.98},
                enlargelimits=0.15,
                legend style={at={(0.5,-0.15)},
                  anchor=north,legend columns=-1},
                ylabel={Macro F1-score},
                symbolic x coords={Chinese,Spanish,English,Russian,Japanese,French,German,Italian},
                xtick=data,
                ]
            \addplot coordinates {(Chinese,0.923)(Spanish,0.943) (English,0.967) (Russian,0.883)(Japanese,0.922)(French,0.971)(German,0.978)(Italian,0.913)};
            \addplot coordinates {(Chinese,0.923) (Spanish,0.929) (English,0.970) (Russian,0.886)(Japanese,0.916)(French,0.962)(German,0.968)(Italian,0.905)};
            \legend{BERT-base-cased,mBERT-base-cased}
        \end{axis}
    \end{tikzpicture}
    \caption{Macro F1-score for each of the eight languages in MLDoc when trained on size 10,000, for both BERT-base-cased and mBERT-base-cased.\protect\footnotemark}
\end{figure*}
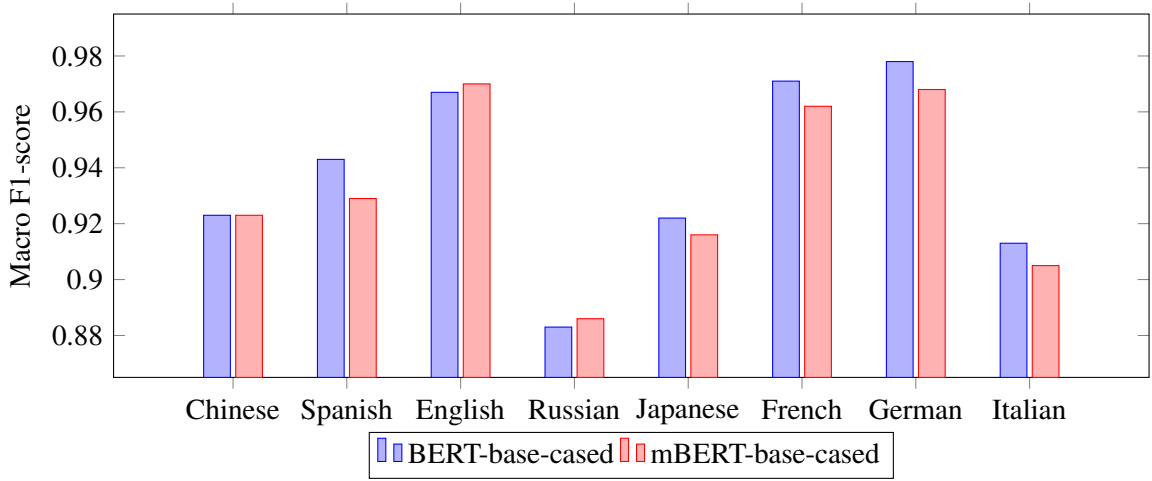

\section{Multilingual Disparity in Text Classification}

We train eight models using texts in each of eight languages, and compare their performance on the task of multi-class text classification. First, we control for the amount of training data across all languages so that any difference in performance can be attributed to either model design or linguistic factors. We then vary the amounts of data used to train these models to observe the possible effects of minimal training data. Additionally, by using the MLDoc corpus we ensure that the distribution of labels in all datasets is identical.

We analyze performance based on the accuracy and F1-score. Results for these metrics, in addition to precision and recall, are displayed in the Appendix. 

\subsection{Text Classification Models}

We investigate eight commonly used text classification models, using default hyperparameters given by open-source libraries: Linear Logistic Regression (LLR) \cite{Bishop:2006:PRM:1162264}, Multinomial Naive Bayes (MNB) \cite{domingos-pedro-pazzani-michael}, and Linear Support Vector Machines (linearSVC) \cite{Fan:2008:LLL:1390681.1442794} models from scikit-learn, Long Short Term Memory (LSTM) \cite{hochreiter-sepp-schmidhuber-jurgen}, Bi-directional Long Short Term Memory (BiLSTM) \cite{hochreiter-sepp-schmidhuber-jurgen}, and Convolutional Neural Network (CNN) \cite{dumoulin2016guide} models from Keras.

\footnotetext{With the exception of Spanish (9458 documents) and Russian (5216 documents), the maximum dataset sizes provided by MLDoc.}

We also use both BERT and Multilingual BERT (mBERT) \cite{devlin-etal-2019-bert}. For BERT, we use a pre-trained BERT model specific to the language at hand. We use BERT for English and Chinese, Italian BERT \footnote{https://github.com/dbmdz/berts\#italian-bert}, German BERT \footnote{https://github.com/dbmdz/berts\#german-bert}, FlauBERT for French \cite{le2019flaubert}, BETO for Spanish \cite{CaneteCFP2020}, RuBERT for Russian \cite{DBLP:journals/corr/abs-1905-07213}, and bert-base-japanese \footnote{https://github.com/cl-tohoku/bert-japanese} for Japanese. All models are fine-tuned for the task of text classification using our training data, with a batch size of 16, maximum sequence length of 250, and learning rate of 2e-5 as recommended by the related work \cite{devlin-etal-2019-bert} for 1 epoch. This allows us to expand the scope of data for language performance differences on a wide range of models.

\subsection{Corpus}

These models are evaluated on a multilingual subset of the Reuters RCV2, MLDoc \cite{SCHWENK18.658}. The Reuters corpus contains news stories in English, German, Spanish, Italian, French, Russian, Chinese, and Japanese. They are sorted into four groups based on the primary subject of each story: CCAT (Corporate/Industrial), ECAT (Economics), GCAT (Government/Social), and MCAT (Markets) \cite{Lewis:2004:RNB:1005332.1005345}. For each language, the data is split into multiple training sets (1000, 5000, and 10000 news stories)\footnotemark[3], a development set (1000 news stories), and a test set (4000 news stories), which MLDoc guarantees uniform class distributions for. Thus, we will analyze model performance on the task of categorizing text samples into one of the four classes for each language.

Admittedly, there are disadvantages to using MLDoc: the corpus is not parallel and nearly 10\% of articles are duplicates \cite{eriksson-2016-quality}; additionally, the languages included are not representative of maximal typological diversity. However, we justify the use of MLDoc as follows. The corpus is representative of a real-world application of NLP technologies. The duplicates noted by \newcite{eriksson-2016-quality} are distributed randomly and thus will not affect the results for any single language. Lastly, although it may be possible to use a dataset that is perhaps more representative of the linguistic variety possible, such as Wikipedia, that would sacrifice either the balance or presence of labels provided by MLDoc.

\begin{figure}[t!]
    \begin{minipage}{0.45\textwidth}
        \begin{tikzpicture}
            \begin{axis}[
                xlabel={\small{Training dataset size}},
                ylabel={\small{Average F1-score}},
                width=0.90\textwidth,
                height=0.27\textheight,
                symbolic x coords={1000, 1002, 3000, 5000, 5216, 9458, 10000},
                xtick = {1000, 5000, 10000},
                xticklabel style={text height=2ex},
                legend pos=outer north east,
                ymajorgrids=true,
                grid style=dashed,
                mark options=solid,
            ]
            \addplot[
                color=red,
                mark=square*,
               ]
                coordinates {
             (1000,0.895)(5000,0.917)(10000,0.923)
                };
            \addlegendentry{\small{C}}
             \addplot[
                color=darkgreen,
                mark=*,
                ]
                coordinates {
             (1000,0.834)(5000,0.895)(10000,0.970)
                };
            \addlegendentry{\small{E}}
             \addplot[
                color=orange,
                mark=triangle*,
                ]
                coordinates {
             (1000,0.855)(5000,0.951)(10000,0.962)
                };
            \addlegendentry{\small{F}}
             \addplot[
                color=black,
                mark=diamond*,
                ]
                coordinates {
             (1000,0.926)(5000,0.961)(10000,0.968)
                };
            \addlegendentry{\small{G}}
             \addplot[
                dashed,
                color=cyan,
                mark=square*,
                ]
                coordinates {
             (1000,0.742)(5000,0.909)(10000,0.905)
                };
            \addlegendentry{\small{I}}
             \addplot[
                dashed,
                color=brown,
                mark=*,
                ]
                coordinates {
             (1000,0.780)(5000,0.900)(10000,0.916)
                };
            \addlegendentry{\small{J}}
             \addplot[
                dashed,
                color=purple,
                mark=triangle*,
                ]
                coordinates {
             (1000,0.797)(5000,0.879)(5216,0.886)
                };
            \addlegendentry{\small{R}}
             \addplot[
                dashed,
                color=blue,
                mark=diamond*,
                ]
                coordinates {
             (1000,0.830)(5000,0.926)(9458,0.929)
                };
            \addlegendentry{\small{S}}
            \end{axis}
        \end{tikzpicture}
        \caption{Comparison between average F1-score with mBERT, on training sizes 1000, 5000, and 10,000 for each language in the MLDoc dataset. The first letter of each language has been used to denote it.\protect\footnotemark}
    \end{minipage}
\end{figure}
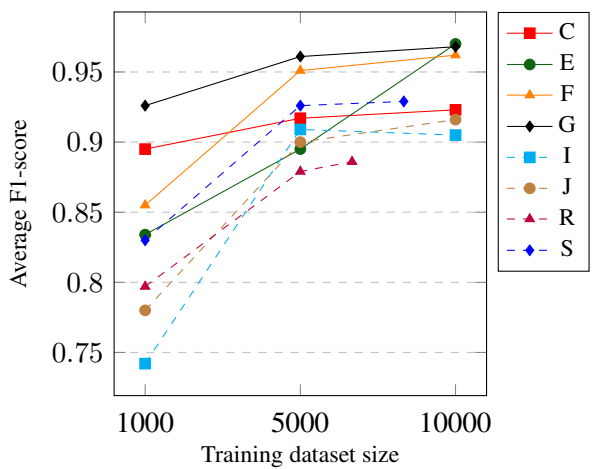

\subsection{Results}

Most models display similar trends in their unequal performance based on language, where models trained on German, English, French, and Spanish consistently outperform models trained on Italian, Russian, Chinese, and Japanese (see Appendix). The highest performing language across the models we tested is German, closely followed by English. Spanish, French, and Italian follow closely, at times outperforming English and German. Inconsistent with Russian's high similarity index (see Table 2), our results show that Russian is the lowest performing European language. This may be because it is morphologically distant from English (see Table 1); in addition, it has flexible word order (second only to German) and high morphological complexity. Another contributing factor could be that the MLDoc dataset provides a maximum training size of 5216 for Russian, in contrast to the 9000 to 10000 provided for all other languages. As illustrated in Figure 2, languages appear to learn at varying rates; for example, the F-1 score for English appears to increase by a similar amount going from 1000 to 5000 versus when going from 5000 to 10000. On the other hand, Spanish increases at a faster rate going from 1000 to 5000, but nearly plateaus in the segment going from 5000 to 10000. If Russian were to follow a similar learning pattern to English, it would be likely that the model's underperformance on Russian could be attributed to insufficient data. 
\footnotetext{With the exception of Spanish (9,458 documents) and Russian (5,216 documents), the maximum dataset sizes provided by MLDoc.}

We note that BERT displays higher scores than mBERT when measured by macro F1-score for every language except English and Russian (for which the differences are negligible: 0.03 for both). However, it is interesting to note that the largest increases in F1-score from mBERT to BERT occur in Spanish, German, French, and Italian (see Figure 1). This indicates that these languages gain the most performance when in use by a model that has been pretrained only by the same language, and thus benefit the least from the use of mBERT, which has been trained on multiple languages. 

For the non-BERT models, training on Chinese largely produces the least effective results, with the margin of difference between the accuracy of Chinese and the top-performing language generally being between 20 to 30 percent for most models. Japanese is not significantly better, and at times under performs Chinese, especially when measured by precision. The macro and weighted recall and F1-scores for Chinese are often below 70$\%$, which is rarely the case with any of the European languages. The highest precision, recall, and F1-scores always correspond to German, French, and Spanish (see Appendix for further details). 

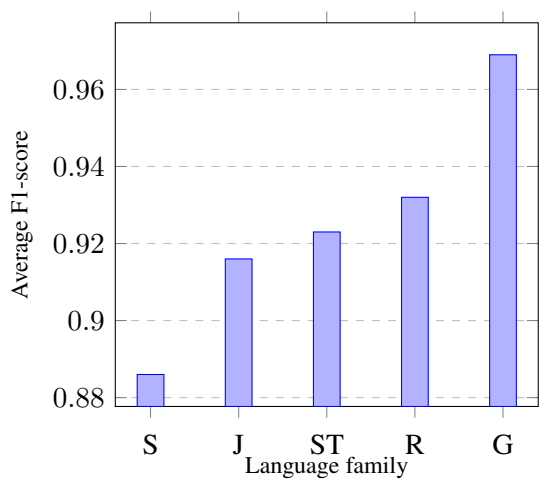
\begin{figure}
    \begin{minipage}[b]{0.47\textwidth}
        \begin{tikzpicture}
            \begin{axis}[
                width=0.95\textwidth,
                height=0.27\textheight,
                 ybar,
                xlabel={\small{Language family}},
                ylabel={\small{Average F1-score}},
                symbolic x coords={S, J, ST, R, G},
                xtick = {S, J, ST, R, G},
                xticklabel style={text height=2ex},
                legend pos=north west,
                ymajorgrids=true,
                grid style=dashed,
            ]
            \addplot coordinates { 
                (S,0.886)(J,0.916)(ST,0.923)(R,0.932)(G,0.969)};
            \end{axis}
        \end{tikzpicture}
        \caption{Average F1-score with mBERT, on training size 10,000, of each language family in the MLDoc dataset. The x-axis corresponds to Slavic, Japonic, Sino-Tibetan, Romance, and Germanic.\footnotemark[7]}
    \end{minipage}
\end{figure}

\section{Analysis}

For each of the eight models with which we tested our languages, we now take the average F1-score on a training size of 10,000. The top four performing languages across our models were German, English, French, and Spanish, which we classify as high-performing languages. Thus, we will consider the remaining four -- Italian, Russian, Chinese, and Japanese -- as low-performing languages. We also include discussion of comparative linguistics within our analysis, for which Figure 3 provides a useful guide.

\subsection{Effects of Resource Availability}

We note that there is a generally positive trend for all languages' performance when trained on increasingly large datasets, as can be seen in Figure 2. Although this is not a new observation, it is a noteworthy one, as it indicates that because larger training sizes improve modeling efficacy, it is worthwhile for future work to expand datasets from predominately underrepresented languages.

We expect languages with lower initial performance to increase more rapidly when given larger training sizes. For instance, it is intuitive that Spanish improves more than German when moving from training size 1000 to 5000 (8.2\% versus 3.5\% with mBERT), since Spanish has a lower initial value than German (85.1\% and 92.6\% with mBERT, respectively). However, there are exceptions to this trend. We note that Italian, Russian, Japanese, French, and Spanish increase at faster rates than English, German, and Chinese, and that these distinctions are not in line with our categorizations of high and low-performing languages.

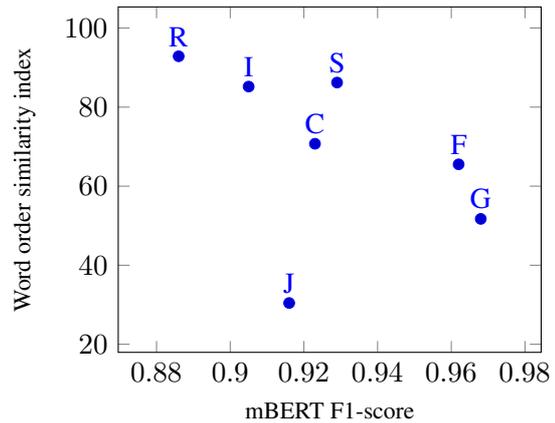
\begin{figure}
    \begin{minipage}[b]{0.47\textwidth}
        \begin{tikzpicture}
            \begin{axis}[
                width=0.95\textwidth,
                enlargelimits=0.2,
                xlabel={\small{mBERT F1-score}},
                ylabel={\small{Word order similarity index}}
                ]
                \addplot+[nodes near coords,only marks,
                point meta=explicit symbolic]
                table[meta=label] {
                x y label
                0.923 70.73 C
                0.962 65.52 F
                0.968 51.72 G
                0.905 85.19 I
                0.916 30.44 J
                0.886 92.86 R
                0.929 86.21 S
                };
            \end{axis}
        \end{tikzpicture}
        \caption{Correlation between the word order similarity index and average F1-scores on mBERT for each language in MLDoc on training size 10,000. The first letter of each language has been used.}
    \end{minipage}
\end{figure}

\subsection{Features of High-Performing Languages}

Overall, it is surprising that German is the highest performing language of the MLDoc dataset, given that it is dissimilar to English in our primary orders of typology: constituent order, word order flexibility, and morphological categorization. Whereas English is a rigid SVO language, German is flexible and is categorized as both SVO and SOV; in addition, English is an analytic language, whereas German is synthetic. It is also important to note that German is consistently the highest performing language across all the models of our study, despite their significantly different architectures (although it is occasionally surpassed by French, Spanish, or English, and then only by a small margin).

Of the other consistently high-performing languages (French, Spanish, and English), typological properties are much more consistent. All three have an SVO constituent order. Although French and Spanish are synthetic fusional languages, while English is analytic, English still contains a moderate amount of inflection and thus there is not a substantial amount of morphological difference between the three. 

We noted previously from a comparative linguistics standpoint that although English and German are both from the Germanic sub-family, Modern English more closely resembles French, because of its use of loanwords and grammatical patterns from Old French (see Section 3.3). Thus it is not entirely unexpected that English, French, and Spanish have roughly equivalent performance results. What is more difficult to account for is the high performance of German, despite its relative typological irregularity and morphological complexity. 

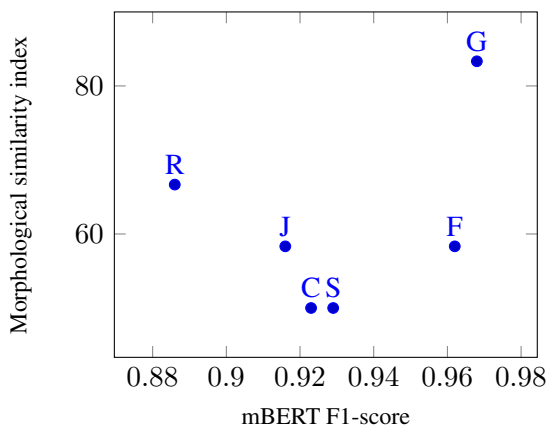
\begin{figure}[t!]
    \begin{minipage}[b]{0.47\textwidth}
        \begin{tikzpicture}
            \begin{axis}[
                width=0.95\textwidth,
                enlargelimits=0.2,
                xlabel={\small{mBERT F1-score}},
                ylabel={\small{Morphological similarity index}}
                ]
                \addplot+[nodes near coords,only marks,
                point meta=explicit symbolic]
                table[meta=label] {
                x y label
                0.923 50 C
                0.962 58.33 F
                0.968 83.33 G
                0.916 58.33 J
                0.886 66.67 R
                0.929 50 S
                };
            \end{axis}
        \end{tikzpicture}
        \caption{Correlation between the morphological similarity index and average F1-scores on mBERT for each language in MLDoc on training size 10,000. The first letter of each language has been used. \protect\footnotemark}
    \end{minipage}
\end{figure}

\subsection{Features of Low-Performing Languages}

Although we mentioned that Italian, Russian, Chinese, and Japanese are the lower four performing languages, this is not without some interesting caveats. 

First, it is important to note that Italian falls under our category of ``low-performing" because our modeling results with Italian are lower than the four languages discussed in Section 6.1. However, the difference between Italian and the ``high-performing" languages is smaller than the difference between Italian and the other languages of this section: Russian, Chinese, and Japanese. This is not particularly surprising, as Italian has an SVO constituent order, a fusional morphological structure, and historically has developed under similar circumstances to our other Romance languages, French and Spanish. 

Next, Chinese has different performance results depending on the model that uses it: for the majority of our models, it is the lowest-performing language, yet for BERT its margin of difference with the top-performing language is 2.9\%, and 3.3\% for mBERT (both on training sizes of 10,000). Additionally, for BERT and mBERT, the top-performing language is English rather than German; and as can be seen from Table 1, English and Chinese have similar syntactic and morphological structures. Thus, we conclude that BERT's architecture is more responsive to typological features than the other models we have included in this study. In a similar matter, we conclude that since linguistic features seem to be weighted less in our non-BERT models, these models are more dependent on other language-specific elements; for example, written Chinese and Japanese are composed of logograms, while the remaining languages utilize phonemic writing.

\footnotetext{Italian is not included in Figure 5; see Section 4.}

\subsection{Similarity Index Correlations}

When comparing word order similarity indices and morphological similarities against our F1-score data for mBERT, we find inconclusive results. Word order similarity seems to have an inverse correlation with modeling efficacy, with Japanese as an obvious outlier (see Figure 4). Morphological similarity does not have a discernable correlation (see Figure 5).

This may be because some of our indices seem counterintuitive. For instance, given that both English and German are Germanic languages, we may consider them to be similar in a general sense. However, we would hardly expect a high similarity index in a morphological context. Consulting Table 1, we see that English is an analytic language, while German is synthetic and fusional; and English is the most rigid language of our dataset, while German is the most flexible (as we have noted, increased word order flexibility is a good indication of high morphological complexity). Additionally, Figure 4 displays an inverse correlation between word order similarity and mBERT, which is also illogical.

Looking into the specific features that we include as factors in the morphological similarity index, we hypothesize that certain morphological attributes might hold greater importance to language similarity than others (our formula for similarity indices placed equal weight on each WALS feature). For example, we find that for feature 29A, Syncretism in Verbal Person/Number Marking, English and low-performing languages are dissimilar whereas the high-performing languages are similar \cite{wals-29}. In future works, it may be good to look into such sub-attributes. 

We attribute much of this to the features available in WALS. However, as is visible in Figure 3, language family is still indicative of model performance. Because languages from the same family are more likely to have derived similar typological characteristics, we conclude that linguistic factors still largely determine modeling efficacy.

\section{Conclusion}

We have provided a thorough investigation of linguistics from an NLP perspective and demonstrated that the variance in language modeling efficacy can be attributed to typological differences in the languages modeled. We have also shown that language family is a strong indication of how well a language will perform. Although we show little correlation between word order or morphological similarity and the strength of a language's modeling results, we note that language families share typological attributes, and thus conclude that a more nuanced definition of language similarity is necessary to evaluate this point. It is crucial that developers keep these findings in mind while creating and evaluating NLP tools.

\section*{Acknowledgments}

We thank Dr. Bernard Comrie (University of California, Santa Barbara) for his expertise regarding cross-lingual typology and discussions in the topic, and Anusha Anand (University of South Carolina) for providing us with direction in which linguistic features to research. 

\bibliography{emnlp2020}

\begin{thebibliography}{35}
\expandafter\ifx\csname natexlab\endcsname\relax\def\natexlab#1{#1}\fi

\bibitem[{Baerman and Brown(2013)}]{wals-29}
Matthew Baerman and Dunstan Brown. 2013.
\newblock \href {https://wals.info/chapter/29} {Syncretism in verbal
  person/number marking}.
\newblock In Matthew~S. Dryer and Martin Haspelmath, editors, \emph{The World
  Atlas of Language Structures Online}. Max Planck Institute for Evolutionary
  Anthropology, Leipzig.

\bibitem[{Bender(2009)}]{Bender:2009:LNL:1642038.1642044}
Emily~M. Bender. 2009.
\newblock \href {http://dl.acm.org/citation.cfm?id=1642038.1642044}
  {Linguistically na\"{I}ve != language independent: Why nlp needs linguistic
  typology}.
\newblock In \emph{Proceedings of the EACL 2009 Workshop on the Interaction
  Between Linguistics and Computational Linguistics: Virtuous, Vicious or
  Vacuous?}, ILCL '09, pages 26--32, Stroudsburg, PA, USA. Association for
  Computational Linguistics.

\bibitem[{Bishop(2006)}]{Bishop:2006:PRM:1162264}
Christopher~M. Bishop. 2006.
\newblock \emph{Pattern Recognition and Machine Learning (Information Science
  and Statistics)}.
\newblock Springer-Verlag, Berlin, Heidelberg.

\bibitem[{Bozsahin(2020)}]{Bozsahin}
Cem Bozsahin. 2020.
\newblock Word order, word order flexibility and the lexicon.

\bibitem[{Bybee(1997)}]{bybee_1997}
Joan~L. Bybee. 1997.
\newblock Semantic aspects of morphological typology.
\newblock In Sandra A.~Thompson Joan L.~Bybee, John~Haiman, editor,
  \emph{Essays on Language Function and Language Type}, pages 25--37. John
  Benjamins Publishing.

\bibitem[{Cañete et~al.(2020)Cañete, Chaperon, Fuentes, and
  Pérez}]{CaneteCFP2020}
José Cañete, Gabriel Chaperon, Rodrigo Fuentes, and Jorge Pérez. 2020.
\newblock Spanish pre-trained bert model and evaluation data.
\newblock In \emph{to appear in PML4DC at ICLR 2020}.

\bibitem[{Comrie(1989)}]{comrie1989language}
B.~Comrie. 1989.
\newblock \href {https://books.google.com/books?id=kf-shTfcaPEC}
  {\emph{Language Universals and Linguistic Typology: Syntax and Morphology}}.
\newblock University of Chicago Press.

\bibitem[{Comrie(2016)}]{comrie_2016}
Bernard Comrie. 2016.
\newblock \href {https://doi.org/10.1075/slcs.173.16com} {\emph{Measuring
  language typicality, with special reference to the Americas}}, pages
  363--384.

\bibitem[{Cotterell et~al.(2018)Cotterell, Mielke, Eisner, and
  Roark}]{cotterell-etal-2018-languages}
Ryan Cotterell, Sebastian~J. Mielke, Jason Eisner, and Brian Roark. 2018.
\newblock \href {https://doi.org/10.18653/v1/N18-2085} {Are all languages
  equally hard to language-model?}
\newblock In \emph{Proceedings of the 2018 Conference of the North {A}merican
  Chapter of the Association for Computational Linguistics: Human Language
  Technologies, Volume 2 (Short Papers)}, pages 536--541, New Orleans,
  Louisiana. Association for Computational Linguistics.

\bibitem[{Cotterell et~al.(2016)Cotterell, Sch{\"u}tze, and
  Eisner}]{cotterell-etal-2016-morphological-smoothing}
Ryan Cotterell, Hinrich Sch{\"u}tze, and Jason Eisner. 2016.
\newblock \href {https://doi.org/10.18653/v1/P16-1156} {Morphological smoothing
  and extrapolation of word embeddings}.
\newblock In \emph{Proceedings of the 54th Annual Meeting of the Association
  for Computational Linguistics (Volume 1: Long Papers)}, pages 1651--1660,
  Berlin, Germany. Association for Computational Linguistics.

\bibitem[{Devlin et~al.(2019)Devlin, Chang, Lee, and
  Toutanova}]{devlin-etal-2019-bert}
Jacob Devlin, Ming-Wei Chang, Kenton Lee, and Kristina Toutanova. 2019.
\newblock \href {https://doi.org/10.18653/v1/N19-1423} {{BERT}: Pre-training of
  deep bidirectional transformers for language understanding}.
\newblock In \emph{Proceedings of the 2019 Conference of the North {A}merican
  Chapter of the Association for Computational Linguistics: Human Language
  Technologies, Volume 1 (Long and Short Papers)}, pages 4171--4186,
  Minneapolis, Minnesota. Association for Computational Linguistics.

\bibitem[{Domingos and Pazzani(1997)}]{domingos-pedro-pazzani-michael}
Pedro Domingos and Michael Pazzani. 1997.
\newblock On the optimality of the simple bayesian classifier under
  zero-oneloss.
\newblock \emph{Machine Learning - ML}, 29:103--130.

\bibitem[{Dryer(2013)}]{wals-81}
Matthew~S. Dryer. 2013.
\newblock \href {https://wals.info/chapter/81} {Order of subject, object and
  verb}.
\newblock In Matthew~S. Dryer and Martin Haspelmath, editors, \emph{The World
  Atlas of Language Structures Online}. Max Planck Institute for Evolutionary
  Anthropology, Leipzig.

\bibitem[{Dryer and Haspelmath(2013)}]{wals}
Matthew~S. Dryer and Martin Haspelmath, editors. 2013.
\newblock \href {https://wals.info/} {\emph{WALS Online}}.
\newblock Max Planck Institute for Evolutionary Anthropology, Leipzig.

\bibitem[{Dumoulin and Visin(2016)}]{dumoulin2016guide}
Vincent Dumoulin and Francesco Visin. 2016.
\newblock \href {http://arxiv.org/abs/1603.07285} {A guide to convolution
  arithmetic for deep learning}.

\bibitem[{Eriksson(2016)}]{eriksson-2016-quality}
Robin Eriksson. 2016.
\newblock \href {https://www.aclweb.org/anthology/L16-1286} {Quality assessment
  of the {R}euters vol. 2 multilingual corpus}.
\newblock In \emph{Proceedings of the Tenth International Conference on
  Language Resources and Evaluation ({LREC} 2016)}, pages 1813--1819,
  Portoro{\v{z}}, Slovenia. European Language Resources Association (ELRA).

\bibitem[{Fan et~al.(2008)Fan, Chang, Hsieh, Wang, and
  Lin}]{Fan:2008:LLL:1390681.1442794}
Rong-En Fan, Kai-Wei Chang, Cho-Jui Hsieh, Xiang-Rui Wang, and Chih-Jen Lin.
  2008.
\newblock \href {http://dl.acm.org/citation.cfm?id=1390681.1442794} {Liblinear:
  A library for large linear classification}.
\newblock \emph{J. Mach. Learn. Res.}, 9:1871--1874.

\bibitem[{Gao(2008)}]{Qian}
Qian Gao. 2008.
\newblock Word order in mandarin: Reading and speaking.
\newblock In \emph{Proceedings of the 20th North American Conference on Chinese
  Linguistics}.

\bibitem[{Garland(2006)}]{Garland_morphologicaltypology}
Jennifer Garland. 2006.
\newblock Morphological typology and the complexity of nominal morphology in
  sinhala.
\newblock In \emph{Proceedings from the Workshop on Sinhala Linguistics}.

\bibitem[{Genetti(2014)}]{genetti2014}
Carol Genetti. 2014.
\newblock \emph{How Languages Word: An Introduction to Languages and
  Linguistics}.
\newblock Cambridge University Press.

\bibitem[{Gordon and Grimes(2019)}]{Ethnologue}
R.~G. Gordon and B.~F. Grimes. 2019.
\newblock \emph{Ethnologue: Languages of the World}, 22nd edition.
\newblock SIL International, Dallas, Texas.

\bibitem[{Hochreiter and
  Schmidhuber(1997)}]{hochreiter-sepp-schmidhuber-jurgen}
Sepp Hochreiter and Jürgen Schmidhuber. 1997.
\newblock \href {https://doi.org/10.1162/neco.1997.9.8.1735} {Long short-term
  memory}.
\newblock \emph{Neural computation}, 9:1735--80.

\bibitem[{Joshi et~al.(2020)Joshi, Santy, Budhiraja, Bali, and
  Choudhury}]{joshi2020state}
Pratik Joshi, Sebastin Santy, Amar Budhiraja, Kalika Bali, and Monojit
  Choudhury. 2020.
\newblock \href {http://arxiv.org/abs/2004.09095} {The state and fate of
  linguistic diversity and inclusion in the nlp world}.
\newblock In \emph{to appear in Proceedings of the 58th Annual Meeting of the
  Association for Computational Linguistics}.

\bibitem[{K et~al.(2020)K, Wang, Mayhew, and Roth}]{k2019crosslingual}
Karthikeyan K, Zihan Wang, Stephen Mayhew, and Dan Roth. 2020.
\newblock \href {https://openreview.net/forum?id=HJeT3yrtDr} {Cross-lingual
  ability of multilingual bert: An empirical study}.
\newblock In \emph{International Conference on Learning Representations}.

\bibitem[{Kuratov and Arkhipov(2019)}]{DBLP:journals/corr/abs-1905-07213}
Yuri Kuratov and Mikhail Arkhipov. 2019.
\newblock \href {http://arxiv.org/abs/1905.07213} {Adaptation of deep
  bidirectional multilingual transformers for russian language}.
\newblock \emph{CoRR}, abs/1905.07213.

\bibitem[{Lahousse and Lamiroy(2012)}]{Lahousse}
Karen Lahousse and Béatrice Lamiroy. 2012.
\newblock \href {https://doi.org/10.1515/flin.2012.014} {Word order in french,
  spanish and italian: A grammaticalization account}.
\newblock \emph{Folia Linguistica}, 2:387--415.

\bibitem[{Le et~al.(2019)Le, Vial, Frej, Segonne, Coavoux, Lecouteux, Allauzen,
  Crabbé, Besacier, and Schwab}]{le2019flaubert}
Hang Le, Loïc Vial, Jibril Frej, Vincent Segonne, Maximin Coavoux, Benjamin
  Lecouteux, Alexandre Allauzen, Benoît Crabbé, Laurent Besacier, and Didier
  Schwab. 2019.
\newblock \href {http://arxiv.org/abs/1912.05372} {Flaubert: Unsupervised
  language model pre-training for french}.

\bibitem[{Lewis et~al.(2004)Lewis, Yang, Rose, and
  Li}]{Lewis:2004:RNB:1005332.1005345}
David~D. Lewis, Yiming Yang, Tony~G. Rose, and Fan Li. 2004.
\newblock \href {http://dl.acm.org/citation.cfm?id=1005332.1005345} {Rcv1: A
  new benchmark collection for text categorization research}.
\newblock \emph{J. Mach. Learn. Res.}, 5:361--397.

\bibitem[{Luong et~al.(2013)Luong, Socher, and
  Manning}]{luong-etal-2013-better}
Thang Luong, Richard Socher, and Christopher Manning. 2013.
\newblock \href {https://www.aclweb.org/anthology/W13-3512} {Better word
  representations with recursive neural networks for morphology}.
\newblock In \emph{Proceedings of the Seventeenth Conference on Computational
  Natural Language Learning}, pages 104--113, Sofia, Bulgaria. Association for
  Computational Linguistics.

\bibitem[{Mielke et~al.(2019)Mielke, Cotterell, Gorman, Roark, and
  Eisner}]{mielke-etal-2019-kind}
Sebastian~J. Mielke, Ryan Cotterell, Kyle Gorman, Brian Roark, and Jason
  Eisner. 2019.
\newblock \href {https://doi.org/10.18653/v1/P19-1491} {What kind of language
  is hard to language-model?}
\newblock In \emph{Proceedings of the 57th Annual Meeting of the Association
  for Computational Linguistics}, pages 4975--4989, Florence, Italy.
  Association for Computational Linguistics.

\bibitem[{Mikolov et~al.(2013)Mikolov, Chen, Corrado, and
  Dean}]{mikolov2013efficient}
Tomas Mikolov, Kai Chen, Gregory~S. Corrado, and Jeffrey Dean. 2013.
\newblock Efficient estimation of word representations in vector space.
\newblock \emph{CoRR}, abs/1301.3781.

\bibitem[{Pennington et~al.(2014)Pennington, Socher, and
  Manning}]{pennington-etal-2014-glove}
Jeffrey Pennington, Richard Socher, and Christopher Manning. 2014.
\newblock \href {https://doi.org/10.3115/v1/D14-1162} {{G}love: Global vectors
  for word representation}.
\newblock In \emph{Proceedings of the 2014 Conference on Empirical Methods in
  Natural Language Processing ({EMNLP})}, pages 1532--1543, Doha, Qatar.
  Association for Computational Linguistics.

\bibitem[{Pires et~al.(2019)Pires, Schlinger, and
  Garrette}]{pires2019multilingual}
Telmo Pires, Eva Schlinger, and Dan Garrette. 2019.
\newblock \href {https://doi.org/10.18653/v1/P19-1493} {How multilingual is
  multilingual {BERT}?}
\newblock In \emph{Proceedings of the 57th Annual Meeting of the Association
  for Computational Linguistics}, pages 4996--5001, Florence, Italy.
  Association for Computational Linguistics.

\bibitem[{Sakel(2015)}]{59036}
Jeanette Sakel. 2015.
\newblock \emph{Study skills for linguistics}.
\newblock Understanding language series.

\bibitem[{Schwenk and Li(2018)}]{SCHWENK18.658}
Holger Schwenk and Xian Li. 2018.
\newblock A corpus for multilingual document classification in eight languages.
\newblock In \emph{Proceedings of the Eleventh International Conference on
  Language Resources and Evaluation (LREC 2018)}, Paris, France. European
  Language Resources Association (ELRA).

\end{thebibliography}
\bibliographystyle{acl_natbib}

\appendix

\onecolumn
\section*{Appendix}

\begin{table*}[hbt!]
    \small
    \begin{center}
        \begin{tabular}{| m{2cm} | m{2cm} | m{2cm} | m{2cm} | m{2cm} | m{2cm} |} 

            \hline
            \multicolumn{6}{|c|}{\textit{Linear Logistic Regression}} \\ \hline
            \hline
            \hfil \centering{Training Size} & \hfil Languages & \hfil Accuracy & 
            \hfil \begin{tabular}{@{}c@{}}Precision \\ \scriptsize{(macro, weighted)}\end{tabular}  & 
            \hfil \begin{tabular}{@{}c@{}}Recall \\ \scriptsize{(macro, weighted)}\end{tabular} & 
            \hfil \begin{tabular}{@{}c@{}}F1-Score \\ \scriptsize{(macro, weighted)}\end{tabular} \\
            
            \hline
             & \hfil { Chinese} & \hfil { 0.678} & \hfil 0.737, 0.739 & \hfil 0.565, 0.678 & \hfil 0.549, 0.678\\ 
             & \hfil { English} & \hfil { 0.899} & \hfil 0.900, 0.901 & \hfil 0.899, 0.899 & \hfil 0.899, 0.899 \\ 
             & \hfil { French} & \hfil { 0.899} & \hfil 0.899, 0.899 & \hfil 0.899, 0.899 & \hfil 0.899, 0.899 \\
            \hfil 1000 & \hfil { German} & \hfil 0.907 & \hfil 0.909, 0.908 & \hfil  0.908, 0.907 & \hfil 0.908, 0.907 \\
             & \hfil { Italian} & \hfil { 0.851} & \hfil 0.854, 0.853 & \hfil 0.852, 0.851 & \hfil 0.851, 0.851 \\
             & \hfil { Japanese} & \hfil 0.682 & \hfil 0.745, 0.748 & \hfil 0.680, 0.682 & \hfil 0.683, 0.682 \\
             & \hfil { Russian} & \hfil { 0.835} & \hfil 0.834, 0.836 & \hfil 0.835, 0.835 & \hfil 0.834, 0.835 \\
             & \hfil { Spanish} & \hfil { 0.925} & \hfil 0.918, 0.926 & \hfil 0.919, 0.925 & \hfil 0.918, 0.925 \\
            \hline
            
             & \hfil { Chinese} & \hfil { 0.736} & \hfil 0.824, 0.789 & \hfil 0.613, 0.736 & \hfil 0.593, 0.736 \\ 
             & \hfil { English} & \hfil { 0.931} & \hfil 0.931, 0.931 & \hfil 0.931, 0.931 & \hfil 0.931, 0.931 \\ 
             & \hfil { French} & \hfil { 0.939} & \hfil 0.939, 0.939 & \hfil 0.939, 0.939 & \hfil 0.939, 0.939 \\
            \hfil 5000 & \hfil { German} & \hfil { 0.937} & \hfil 0.938, 0.937 & \hfil 0.937, 0.937 & \hfil 0.938, 0.937 \\
             & \hfil { Italian} & \hfil { 0.892} & \hfil 0.893, 0.892 & \hfil 0.892, 0.892 & \hfil 0.892, 0.892 \\
             & \hfil { Japanese} & \hfil { 0.747} & \hfil 0.782, 0.785 & \hfil 0.746, 0.747 & \hfil 0.749, 0.746 \\
             & \hfil { Russian} & \hfil { 0.836} & \hfil 0.849, 0.843 & \hfil 0.821, 0.836 & \hfil 0.829, 0.83 \\
             & \hfil { Spanish} & \hfil { 0.899} & \hfil 0.907, 0.902 & \hfil 0.880, 0.899 & \hfil 0.890, 0.899 \\
             
             \hline 
             \hfil 5216 & \hfil Russian & \hfil { 0.836} & \hfil 0.848, 0.842 & \hfil 0.821, 0.836 & \hfil 0.829, 0.836 \\
             
             \hline 
             \hfil 9458 & \hfil Spanish & \hfil { 0.906} & \hfil 0.917, 0.911 & \hfil 0.890, 0.906 & \hfil 0.899, 0.906 \\
             
             \hline
             & \hfil { Chinese} & \hfil { 0.768} & \hfil 0.834, 0.802 & \hfil 0.635, 0.768 & \hfil 0.609, 0.768 \\ 
             & \hfil { English} & \hfil { 0.935} & \hfil 0.935, 0.935 & \hfil 0.934, 0.935 & \hfil 0.934, 0.935 \\ 
             & \hfil { French} & \hfil { 0.945} & \hfil 0.945, 0.945 & \hfil 0.944, 0.945 & \hfil 0.944, 0.945 \\
            \hfil 10000 & \hfil { German} & \hfil { 0.929} & \hfil 0.931, 0.929 & \hfil 0.929, 0.929 & \hfil 0.929, 0.929 \\
             & \hfil { Italian} & \hfil { 0.849} & \hfil 0.858, 0.858 & \hfil 0.849, 0.849 & \hfil 0.848, 0.849 \\
             & \hfil { Japanese} & \hfil { 0.774} & \hfil 0.794, 0.797 & \hfil 0.773, 0.774 & \hfil 0.776, 0.774 \\
             \hline
        \end{tabular}
    \end{center}
        
    \begin{center}
        \begin{tabular}{| m{2cm} | m{2cm} | m{2cm} | m{2cm} | m{2cm} | m{2cm} |} 

            \hline
            \multicolumn{6}{|c|}{\textit{Multinomial Naive Bayes}} \\ \hline
            \hline
            \hfil \centering{Training Size} & \hfil Languages & \hfil Accuracy & 
            \hfil \begin{tabular}{@{}c@{}}Precision \\ \scriptsize{(macro, weighted)}\end{tabular}  & 
            \hfil \begin{tabular}{@{}c@{}}Recall \\ \scriptsize{(macro, weighted)}\end{tabular} & 
            \hfil \begin{tabular}{@{}c@{}}F1-Score \\ \scriptsize{(macro, weighted)}\end{tabular} \\ 
            
            \hline
             & \hfil Chinese & \hfil 0.727 & \hfil 0.797, 0.755 & \hfil 0.600, 0.727 & \hfil 0.573, 0.727 \\ 
             & \hfil English & \hfil 0.902 & \hfil 0.903, 0.903 & \hfil 0.902, 0.902 & \hfil 0.902, 0.902 \\ 
             & \hfil French & \hfil 0.902 & \hfil 0.905, 0.905 & \hfil 0.902, 0.902 & \hfil 0.902, 0.902 \\
            \hfil 1000 & \hfil German & \hfil 0.920 & \hfil 0.922, 0.922 & \hfil 0.921, 0.920 & \hfil 0.919, 0.920 \\
             & \hfil Italian & \hfil 0.844 & \hfil 0.845, 0.844 & \hfil 0.846, 0.844 & \hfil 0.844, 0.844 \\
             & \hfil Japanese & \hfil 0.736 & \hfil 0.735, 0.737 & \hfil 0.733, 0.736 & \hfil 0.732, 0.736 \\
             & \hfil Russian & \hfil 0.753 & \hfil 0.749, 0.763 & \hfil 0.733, 0.753 & \hfil 0.737, 0.753 \\
             & \hfil Spanish & \hfil 0.925 & \hfil 0.929, 0.927 & \hfil 0.910, 0.925 & \hfil 0.918, 0.925 \\
             
            \hline
             & \hfil { Chinese} & \hfil { 0.776} & \hfil 0.647, 0.738 & \hfil 0.642, 0.776 & \hfil 0.615, 0.776 \\ 
             & \hfil { English} & \hfil { 0.922} & \hfil 0.923, 0.923 & \hfil 0.921, 0.922 & \hfil 0.921, 0.922 \\ 
             & \hfil { French} & \hfil { 0.933} & \hfil 0.933, 0.933 & \hfil 0.932, 0.933 & \hfil 0.932, 0.933 \\
            \hfil 5000 & \hfil { German} & \hfil 0.955 & \hfil 0.956, 0.955 & \hfil 0.955, 0.955 & \hfil 0.955, 0.955 \\
             & \hfil { Italian} & \hfil { 0.862} & \hfil 0.866, 0.864 & \hfil 0.862, 0.862 & \hfil 0.862, 0.862 \\
             & \hfil { Japanese} & \hfil { 0.798} & \hfil 0.806, 0.809 & \hfil 0.798, 0.798 & \hfil 0.799, 0.798 \\
             & \hfil { Russian} & \hfil { 0.781} & \hfil 0.777, 0.787 & \hfil 0.759, 0.781 & \hfil 0.764, 0.781 \\
             & \hfil { Spanish} & \hfil { 0.899} & \hfil 0.918, 0.907 & \hfil 0.872, 0.899 & \hfil 0.884, 0.899 \\
             
             \hline 
             \hfil 5216 & \hfil Russian & \hfil { 0.776} & \hfil 0.787, 0.760 & \hfil 0.781, 0.781 & \hfil 0.764, 0.781 \\
             
             \hline 
             \hfil 9458 & \hfil Spanish & \hfil { 0.904} & \hfil 0.921, 0.911 & \hfil 0.880, 0.904 & \hfil 0.891, 0.904 \\
             
             \hline
             & \hfil { Chinese} & \hfil { 0.789} & \hfil 0.657, 0.745 & \hfil 0.651, 0.789 & \hfil 0.621, 0.789 \\ 
             & \hfil { English} & \hfil { 0.927} & \hfil 0.928, 0.928 & \hfil 0.926, 0.927 & \hfil 0.927, 0.927 \\ 
             & \hfil { French} & \hfil { 0.937} & \hfil 0.937, 0.937 & \hfil 0.936, 0.937 & \hfil 0.937, 0.937 \\
            \hfil 10000 & \hfil { German} & \hfil { 0.959} & \hfil 0.959, 0.959 & \hfil 0.959, 0.959 & \hfil 0.959, 0.959 \\
             & \hfil { Italian} & \hfil { 0.862} & \hfil 0.866, 0.865 & \hfil 0.863, 0.862 & \hfil 0.863, 0.862 \\
             & \hfil { Japanese} & \hfil { 0.815} & \hfil 0.823, 0.826 & \hfil 0.815, 0.815 & \hfil 0.816, 0.815 \\
             \hline
        \end{tabular}
    \end{center}
\end{table*}
    
\begin{table*}[hbt!]
    \small
    \begin{center}
        \begin{tabular}{| m{2cm} | m{2cm} | m{2cm} | m{2cm} | m{2cm} | m{2cm} |} 

            \hline
            \multicolumn{6}{|c|}{\textit{LinearSVC}} \\ \hline
            \hline
            \hfil \centering{Training Size} & \hfil Languages & \hfil Accuracy & 
            \hfil \begin{tabular}{@{}c@{}}Precision \\ \scriptsize{(macro, weighted)}\end{tabular}  & 
            \hfil \begin{tabular}{@{}c@{}}Recall \\ \scriptsize{(macro, weighted)}\end{tabular} & 
            \hfil \begin{tabular}{@{}c@{}}F1-Score \\ \scriptsize{(macro, weighted)}\end{tabular} \\
            
            \hline
             & \hfil { Chinese} & \hfil { 0.682} & \hfil 0.653, 0.709 & \hfil 0.584, 0.682 & \hfil 0.584, 0.682\\ 
             & \hfil { English} & \hfil { 0.895} & \hfil 0.897, 0.897 & \hfil 0.895, 0.895 & \hfil 0.895, 0.895 \\ 
             & \hfil { French} & \hfil { 0.912} & \hfil 0.912, 0.913 & \hfil 0.912, 0.912 & \hfil 0.912, 0.912\\
            \hfil { 1000} & \hfil { German} & \hfil { 0.915} & \hfil 0.916, 0.916 & \hfil 0.916, 0.915 & \hfil 0.916, 0.915\\
             & \hfil { Italian} & \hfil {0.859} & \hfil 0.861, 0.859 & \hfil 0.859, 0.859 & \hfil 0.859, 0.859\\
             & \hfil Japanese & \hfil 0.721 & \hfil 0.747, 0.751 & \hfil 0.719, 0.721 & \hfil 0.723, 0.721\\
             & \hfil Russian & \hfil 0.833 & \hfil 0.832, 0.833 & \hfil 0.831, 0.833 & \hfil 0.831, 0.833\\
             & \hfil Spanish & \hfil 0.932 & \hfil 0.925, 0.934 & \hfil 0.929, 0.932 & \hfil 0.927, 0.932\\
             
            \hline
             & \hfil Chinese & \hfil 0.701 & \hfil 0.753, 0.754 & \hfil 0.569, 0.701 & \hfil 0.566, 0.701\\ 
             & \hfil English & \hfil 0.889 & \hfil 0.889, 0.891 & \hfil 0.889, 0.889 & \hfil 0.889, 0.889\\ 
             & \hfil French & \hfil 0.901 & \hfil 0.902, 0.902 & \hfil 0.901, 0.901 & \hfil 0.901, 0.901\\
            \hfil 5000 & \hfil German & \hfil 0.915 & \hfil 0.916, 0.916 & \hfil 0.916, 0.915 & \hfil 0.916, 0.915\\
             & \hfil Italian & \hfil 0.852 & \hfil 0.851, 0.852 & \hfil 0.847, 0.852 & \hfil 0.848, 0.852\\
             & \hfil Japanese & \hfil 0.726 & \hfil 0.745, 0.747 & \hfil 0.724, 0.726 & \hfil 0.729, 0.726\\
             & \hfil Russian & \hfil 0.818 & \hfil 0.824, 0.819 & \hfil 0.798, 0.818  & \hfil 0.808, 0.818\\
             & \hfil Spanish & \hfil 0.902 & \hfil 0.901, 0.902 & \hfil 0.839, 0.902 & \hfil 0.866, 0.902\\
             
             \hline 
             \hfil 5216 & \hfil Russian & \hfil 0.824 & \hfil 0.827, 0.826 & \hfil 0.804, 0.824 & \hfil 0.813, 0.824\\
             
             \hline 
             \hfil 9458 & \hfil Spanish & \hfil 0.901 & \hfil 0.883, 0.898 & \hfil 0.742, 0.901 & \hfil 0.792, 0.901\\
             
             \hline
             & \hfil Chinese & \hfil 0.736 & \hfil 0.757, 0.759 & \hfil 0.577, 0.736 & \hfil 0.564, 0.736\\ 
             & \hfil English & \hfil 0.881 & \hfil 0.882, 0.882 & \hfil 0.881, 0.881 & \hfil 0.881, 0.881\\ 
             & \hfil French & \hfil 0.906 & \hfil 0.906, 0.906 & \hfil 0.905, 0.906 & \hfil 0.905, 0.906\\
            \hfil 10000 & \hfil German & \hfil 0.918 & \hfil 0.918, 0.918 & \hfil 0.918, 0.918 & \hfil 0.918, 0.918\\
             & \hfil Italian & \hfil 0.849 & \hfil 0.847, 0.850 & \hfil 0.818, 0.849 & \hfil 0.829, 0.849\\
             & \hfil Japanese & \hfil 0.723 & \hfil 0.748, 0.751 & \hfil 0.722, 0.723 & \hfil 0.726, 0.723\\
             \hline
        \end{tabular}
    \end{center}
    
    \begin{center}
        \begin{tabular}{| m{2cm} | m{2cm} | m{2cm} | m{2cm} | m{2cm} | m{2cm} |} 

            \hline
            \multicolumn{6}{|c|}{\textit{LSTM}} \\ \hline
            \hline
            \hfil \centering{Training Size} & \hfil Languages & \hfil Accuracy & 
            \hfil \begin{tabular}{@{}c@{}}Precision \\ \scriptsize{(macro, weighted)}\end{tabular}  & 
            \hfil \begin{tabular}{@{}c@{}}Recall \\ \scriptsize{(macro, weighted)}\end{tabular} & 
            \hfil \begin{tabular}{@{}c@{}}F1-Score \\ \scriptsize{(macro, weighted)}\end{tabular} \\ 
             
             \hline
             & \hfil Chinese & \hfil 0.414 & \hfil 0.519, 0.636 & \hfil 0.350, 0.414 & \hfil 0.272, 0.414\\ 
             & \hfil English & \hfil 0.679 & \hfil 0.672, 0.672 & \hfil 0.678, 0.679 & \hfil 0.673, 0.679\\ 
             & \hfil French & \hfil 0.565 & \hfil 0.563, 0.566 & \hfil 0.562, 0.565 & \hfil 0.562, 0.565\\
            \hfil 1000 & \hfil German & \hfil 0.820 & \hfil 0.829, 0.827 & \hfil 0.821, 0.820 & \hfil 0.820, 0.820\\
              & \hfil Italian & \hfil  0.666 & \hfil 0.672, 0.672 & \hfil 0.668, 0.666 & \hfil 0.667, 0.666 \\
              & \hfil Japanese & \hfil 0.249 & \hfil 0.333, 0.339 & \hfil 0.254, 0.249 & \hfil 0.106, 0.249\\
             & \hfil Russian & \hfil 0.727 & \hfil 0.724, 0.724 & \hfil 0.723, 0.727 & \hfil 0.722, 0.727\\
             & \hfil Spanish & \hfil 0.764 & \hfil 0.828, 0.857 & \hfil 0.759, 0.764 & \hfil 0.752, 0.764\\
             \hline
             
             & \hfil Chinese & \hfil 0.490 & \hfil 0.379, 0.460 & \hfil 0.396, 0.490 & \hfil 0.316, 0.490\\ 
             & \hfil English & \hfil 0.846 & \hfil 0.847, 0.847 & \hfil 0.845, 0.846 & \hfil 0.845, 0.846\\ 
             & \hfil French & \hfil 0.882 & \hfil 0.884, 0.885 & \hfil 0.882, 0.882 & \hfil 0.882, 0.882\\
            \hfil 5000 & \hfil German & \hfil 0.909 & \hfil 0.910, 0.909 & \hfil 0.909, 0.908 & \hfil 0.909, 0.909\\
              & \hfil Italian & \hfil  0.870 & \hfil 0.872, 0.871 & \hfil 0.870, 0.870 & \hfil 0.871, 0.870 \\
              & \hfil Japanese & \hfil  0.251 & \hfil 0.619, 0.622 & \hfil 0.256, 0.251 & \hfil 0.110, 0.251\\
             & \hfil Russian & \hfil 0.787 & \hfil 0.789, 0.789 & \hfil 0.777, 0.787 & 0.781, 0.787\\
             & \hfil Spanish & \hfil 0.848 & \hfil 0.837, 0.851 & \hfil 0.830, 0.848 & \hfil 0.831, 0.848\\

             \hline 
             \hfil 5216 & \hfil Russian & \hfil 0.749 & \hfil 0.758, 0.760 & \hfil 0.735, 0.749 & \hfil 0.740, 0.749\\
             
             \hline 
             \hfil 9458 & \hfil Spanish & \hfil 0.764 & \hfil 0.787, 0.778 & \hfil 0.714, 0.764 & \hfil 0.726, 0.764\\
             
             \hline
             & \hfil Chinese & \hfil 0.469 & \hfil 0.255, 0.313 & \hfil 0.379, 0.466 & \hfil 0.295, 0.469\\ 
             & \hfil English & \hfil 0.865 & \hfil 0.870, 0.871 & \hfil 0.865, 0.865 & \hfil 0.866, 0.865\\ 
             & \hfil French & \hfil 0.879 & \hfil 0.879, 0.879 & \hfil 0.878, 0.879 & \hfil 0.878, 0.879\\
            \hfil 10000 & \hfil German & \hfil 0.922 & \hfil 0.922, 0.922 & \hfil 0.923, 0.922 & \hfil 0.922, 0.922\\
              & \hfil Italian & \hfil  0.746 & \hfil 0.768, 0.765 & \hfil 0.743, 0.746 & \hfil 0.722, 0.746 \\
              & \hfil Japanese & \hfil 0.241 & \hfil 0.255, 0.257 & \hfil 0.253, 0.241 & \hfil 0.105, 0.241\\
             \hline
        \end{tabular}
    \end{center}
\end{table*}
    
\begin{table*}[hbt!]
    \small
    \begin{center}
        \begin{tabular}{| m{2cm} | m{2cm} | m{2cm} | m{2cm} | m{2cm} | m{2cm} |} 

            \hline
            \multicolumn{6}{|c|}{\textit{BiLSTM}} \\ \hline
            \hline
            \hfil \centering{Training Size} & \hfil Languages & \hfil Accuracy & 
            \hfil \begin{tabular}{@{}c@{}}Precision \\ \scriptsize{(macro, weighted)}\end{tabular}  & 
            \hfil \begin{tabular}{@{}c@{}}Recall \\ \scriptsize{(macro, weighted)}\end{tabular} & 
            \hfil \begin{tabular}{@{}c@{}}F1-Score \\ \scriptsize{(macro, weighted)}\end{tabular} \\
            
            \hline
             & \hfil Chinese & \hfil 0.728 & \hfil 0.684, 0.747 & \hfil 0.704, 0.728 & \hfil 0.688, 0.728\\ 
             & \hfil English & \hfil 0.800 & \hfil 0.798, 0.799 & \hfil 0.799, 0.800 & \hfil 0.800, 0.800\\ 
             & \hfil French & \hfil 0.855 & \hfil 0.859, 0.860 & \hfil 0.855, 0.855 & \hfil 0.855, 0.855\\
            \hfil 1000 & \hfil German & \hfil 0.877 & \hfil 0.878, 0.877 & \hfil 0.878, 0.877 & \hfil 0.878, 0.877\\
             & \hfil Italian & \hfil 0.781 & \hfil 0.781, 0.781 & \hfil 0.781, 0.781 & \hfil 0.780, 0.781\\
             & \hfil Japanese & \hfil 0.568 & \hfil 0.693, 0.696 & \hfil 0.563, 0.568 & \hfil 0.525, 0.568\\
             & \hfil Russian & \hfil 0.809 & \hfil 0.803, 0.813 & \hfil 0.809, 0.809 & \hfil 0.805, 0.809\\
             & \hfil Spanish & \hfil 0.901 & \hfil 0.891, 0.903 & \hfil 0.893, 0.901 & \hfil 0.891, 0.901\\
             
            \hline
             & \hfil Chinese & \hfil 0.740 & \hfil 0.616, 0.705 & \hfil 0.612, 0.740 & \hfil 0.589, 0.740\\ 
             & \hfil English & \hfil 0.885 & \hfil 0.889, 0.890 & \hfil 0.885, 0.885 & \hfil 0.885, 0.885\\ 
             & \hfil French & \hfil 0.893 & \hfil 0.893, 0.894 & \hfil 0.892, 0.893 & \hfil 0.892, 0.893\\
            \hfil 5000 & \hfil German & \hfil 0.925 & \hfil 0.927, 0.926 & \hfil 0.926, 0.925 & \hfil 0.926, 0.925\\
             & \hfil Italian & \hfil 0.843 & \hfil 0.854, 0.853 & \hfil 0.842, 0.843 & \hfil 0.843, 0.843\\
             & \hfil Japanese & \hfil 0.623 & \hfil 0.735, 0.738 & \hfil 0.617, 0.623 & \hfil 0.594, 0.623\\
             & \hfil Russian & \hfil 0.825 & \hfil 0.837, 0.832 & \hfil 0.815, 0.826 & \hfil 0.822, 0.826\\
             & \hfil Spanish & \hfil 0.911 & \hfil 0.910, 0.912 & \hfil 0.899, 0.911 & \hfil 0.902, 0.911\\
             
             \hline 
             \hfil 5216 & \hfil Russian & \hfil 0.817 & \hfil 0.821, 0.818 & \hfil 0.806, 0.817 & \hfil 0.811, 0.817\\
             
             \hline 
             \hfil 9458 & \hfil Spanish & \hfil 0.899 & \hfil 0.901, 0.900 & \hfil 0.884, 0.899 & \hfil 0.890, 0.899\\
             
             \hline
             & \hfil Chinese & \hfil 0.824 & \hfil 0.824, 0.824 & \hfil 0.826, 0.824 & \hfil 0.825, 0.824\\ 
             & \hfil English & \hfil 0.922 & \hfil 0.920, 0.922 & \hfil 0.907,0.922 & \hfil 0.913, 0.922\\ 
             & \hfil French & \hfil 0.930 & \hfil 0.927, 0.933 & \hfil 0.926, 0.930 & \hfil 0.925, 0.930\\
            \hfil 10000 & \hfil German & \hfil 0.940 & \hfil 0.940, 0.940 & \hfil 0.940, 0.940 & \hfil 0.940, 0.940\\
             & \hfil Italian & \hfil 0.873 & \hfil 0.880, 0.880 & \hfil 0.872, 0.873 & \hfil 0.873, 0.873\\
             & \hfil Japanese & \hfil 0.631 & \hfil 0.675, 0.682 & \hfil 0.629, 0.631 & \hfil 0.624, 0.631\\
             \hline
        \end{tabular}
    \end{center}

    \begin{center}
        \begin{tabular}{| m{2cm} | m{2cm} | m{2cm} | m{2cm} | m{2cm} | m{2cm} |} 

            \hline
            \multicolumn{6}{|c|}{\textit{CNN}} \\ \hline
            \hline
            \hfil \centering{Training Size} & \hfil Languages & \hfil Accuracy & 
            \hfil \begin{tabular}{@{}c@{}}Precision \\ \scriptsize{(macro, weighted)}\end{tabular}  & 
            \hfil \begin{tabular}{@{}c@{}}Recall \\ \scriptsize{(macro, weighted)}\end{tabular} & 
            \hfil \begin{tabular}{@{}c@{}}F1-Score \\ \scriptsize{(macro, weighted)}\end{tabular} \\
            
            \hline
             & \hfil Chinese & \hfil 0.699 & \hfil 0.666, 0.739 & \hfil 0.687, 0.699 & \hfil 0.662, 0.699\\ 
             & \hfil English & \hfil 0.879 & \hfil 0.880, 0.880 & \hfil 0.879, 0.879 & \hfil 0.879, 0.879\\ 
             & \hfil French & \hfil 0.888 & \hfil 0.892, 0.893 & \hfil 0.887, 0.888 & \hfil 0.888, 0.888\\
            \hfil 1000 & \hfil German & \hfil 0.912 & \hfil 0.912, 0.911 & \hfil 0.912, 0.912 & \hfil 0.912, 0.912\\
             & \hfil Italian & \hfil 0.829 & \hfil 0.829, 0.829 & \hfil 0.829, 0.829 & \hfil 0.829, 0.829\\
             & \hfil Japanese & \hfil 0.604 & \hfil 0.640, 0.646 & \hfil 0.598, 0.604 & \hfil 0.595, 0.604\\
             & \hfil Russian & \hfil 0.818 & \hfil 0.813, 0.823 & \hfil 0.820, 0.818 & \hfil 0.814, 0.818\\
             & \hfil Spanish & \hfil 0.925 & \hfil 0.917, 0.926 & \hfil 0.921, 0.925 & \hfil 0.919, 0.925\\
             
            \hline
             & \hfil Chinese & \hfil 0.729 & \hfil 0.636, 0.711 & \hfil 0.606, 0.729 & \hfil 0.589, 0.729\\ 
             & \hfil English & \hfil 0.926 & \hfil 0.925, 0.926 & \hfil 0.925, 0.926 & \hfil 0.925, 0.926\\ 
             & \hfil French & \hfil 0.939 & \hfil 0.938, 0.939 & \hfil 0.938, 0.939 & \hfil 0.938, 0.939\\
            \hfil 5000 & \hfil German & \hfil 0.948 & \hfil 0.948, 0.948 & \hfil 0.948, 0.948 & \hfil 0.948, 0.948\\
             & \hfil Italian & \hfil 0.887 & \hfil 0.891, 0.889 & \hfil 0.886, 0.887 & \hfil 0.888, 0.887\\
             & \hfil Japanese & \hfil 0.633 & \hfil 0.733, 0.736 & \hfil 0.626, 0.633 & \hfil 0.601, 0.633\\
             & \hfil Russian & \hfil 0.852 & \hfil 0.856, 0.853 & \hfil 0.845, 0.852 & \hfil 0.849, 0.852\\
             & \hfil Spanish & \hfil 0.928 & \hfil 0.928, 0.928 & \hfil 0.915, 0.928 & \hfil 0.920, 0.928\\
             
             \hline 
             \hfil 5216 & \hfil Russian & \hfil 0.869 & \hfil 0.874, 0.871 & \hfil 0.863, 0.869 & \hfil 0.867, 0.869\\
             
             \hline 
             \hfil 9458 & \hfil Spanish & \hfil 0.916 & \hfil 0.921, 0.919 & \hfil 0.903, 0.916 & \hfil 0.909, 0.916\\
             
             \hline
             & \hfil Chinese & \hfil 0.759 & \hfil 0.576, 0.701 & \hfil 0.626, 0.759 & \hfil 0.598, 0.759\\ 
             & \hfil English & \hfil 0.930 & \hfil 0.931, 0.931 & \hfil 0.930, 0.930 & \hfil 0.930, 0.930\\ 
             & \hfil French & \hfil 0.943 & \hfil 0.943, 0.943 & \hfil 0.942, 0.943 & \hfil 0.942, 0.943\\
            \hfil 10000 & \hfil German & \hfil 0.955 & \hfil 0.955, 0.955 & \hfil 0.955, 0.955 & \hfil 0.955, 0.955\\
             & \hfil Italian & \hfil 0.900 & \hfil 0.900, 0.900 & \hfil 0.900, 0.900 & \hfil 0.900, 0.900\\
             & \hfil Japanese & \hfil 0.658 & \hfil 0.690, 0.700 & \hfil 0.654, 0.659 & \hfil 0.661, 0.658\\
             \hline
        \end{tabular}
    \end{center}
\end{table*}

\begin{table*}[t]
    \small
    \begin{center}
        \begin{tabular}{| m{2cm} | m{2cm} | m{2cm} | m{2cm} | m{2cm} | m{2cm} |} 

            \hline
            \multicolumn{6}{|c|}{\textit{BERT-Base-Cased}} \\ \hline
            \hline
            \hfil \centering{Training Size} & \hfil Languages & \hfil Accuracy & 
            \hfil \begin{tabular}{@{}c@{}}Precision \\ \scriptsize{(macro, weighted)}\end{tabular}  & 
            \hfil \begin{tabular}{@{}c@{}}Recall \\ \scriptsize{(macro, weighted)}\end{tabular} & 
            \hfil \begin{tabular}{@{}c@{}}F1-Score \\ \scriptsize{(macro, weighted)}\end{tabular} \\ 
            
            \hline
            & \hfil Chinese & \hfil 0.881 & \hfil 0.888, 0.885 & \hfil 0.836, 0.881 & \hfil 0.855, 0.880\\ 
             & \hfil English & \hfil 0.876 & \hfil 0.881, 0.881 & \hfil 0.875, 0.876 & \hfil 0.875, 0.876\\
             & \hfil French & \hfil 0.926 & \hfil 0.926, 0.926 & \hfil 0.926, 0.926 & \hfil 0.926, 0.926\\
            \hfil 1000 & \hfil German & \hfil 0.938 & \hfil 0.941, 0.940 & \hfil 0.938, 0.938 & \hfil 0.939, 0.938\\
             & \hfil Italian & \hfil 0.819 & \hfil 0.821, 0.822 & \hfil 0.821, 0.819 & \hfil 0.817, 0.819\\
             & \hfil Japanese & \hfil 0.795 & \hfil 0.796, 0.796 & \hfil 0.794, 0.795 & \hfil 0.793, 0.795\\
             & \hfil Russian & \hfil 0.805 & \hfil 0.798, 0.807 & \hfil 0.792, 0.805 & \hfil 0.794, 0.805\\
             & \hfil Spanish & \hfil 0.916 & \hfil 0.912, 0.918 & \hfil 0.910, 0.916 & \hfil 0.910, 0.916\\
             
             \hline
             & \hfil Chinese & \hfil 0.939 & \hfil 0.942, 0.939 & \hfil 0.915, 0.939 & \hfil 0.926, 0.939\\ 
             & \hfil English & \hfil 0.943 & \hfil 0.944, 0.945 & \hfil 0.943, 0.943 & \hfil 0.943, 0.943\\ 
             & \hfil French & \hfil 0.963 & \hfil 0.963, 0.963 & \hfil 0.963, 0.967 & \hfil 0.963, 0.963\\
            \hfil 5000 & \hfil German & \hfil 0.973 & \hfil 0.974, 0.974 & \hfil 0.973, 0.973 & \hfil 0.973, 0.973\\
             & \hfil Italian & \hfil 0.895 & \hfil 0.900, 0.899 & \hfil 0.895, 0.895 & \hfil 0.896, 0.895\\
             & \hfil Japanese & \hfil 0.910 & \hfil 0.910, 0.911 & \hfil 0.909, 0.910 & \hfil 0.909, 0.910\\
             & \hfil Russian & \hfil 0.882 & \hfil 0.885, 0.884 & \hfil 0.880, 0.882 & \hfil 0.882, 0.882\\
             & \hfil Spanish & \hfil 0.962 & \hfil 0.959, 0.963 & \hfil 0.960, 0.962 & \hfil 0.959, 0.962\\
             
             \hline 
             \hfil 5216 & \hfil Russian & \hfil 0.885 & \hfil 0.890, 0.886 & \hfil 0.878, 0.885 & \hfil 0.883, 0.885\\
             
             \hline 
             \hfil 9458 & \hfil Spanish & \hfil 0.948 & \hfil 0.952, 0.949 & \hfil 0.937, 0.948 & \hfil 0.943, 0.948\\
             
            \hline
            & \hfil Chinese & \hfil 0.936 & \hfil 0.938, 0.937 & \hfil 0.912, 0.936 & \hfil 0.923, 0.936\\ 
             & \hfil English & \hfil 0.967 & \hfil 0.967, 0.967 & \hfil 0.967, 0.967 & \hfil 0.967, 0.967\\ 
             & \hfil French & \hfil 0.971 & \hfil 0.971, 0.971 & \hfil 0.971, 0.971 & \hfil 0.971, 0.971\\
            \hfil 10000 & \hfil German & \hfil 0.978 & \hfil 0.978, 0.978 & \hfil 0.978, 0.978 & \hfil 0.978, 0.978\\
             & \hfil Italian & \hfil 0.912 & \hfil 0.915, 0.914 & \hfil 0.912, 0.912 & \hfil 0.913, 0.912\\
             & \hfil Japanese & \hfil 0.923 & \hfil 0.922, 0.923 & \hfil 0.922, 0.923 & \hfil 0.922, 0.923\\
             
            \hline
        \end{tabular}
    \end{center}
    
    \begin{center}
        \begin{tabular}{| m{2cm} | m{2cm} | m{2cm} | m{2cm} | m{2cm} | m{2cm} |} 

            \hline
            \multicolumn{6}{|c|}{\textit{mBERT-Base-Cased}} \\ \hline
            \hline
            \hfil \centering{Training Size} & \hfil Languages & \hfil Accuracy & 
            \hfil \begin{tabular}{@{}c@{}}Precision \\ \scriptsize{(macro, weighted)}\end{tabular}  & 
            \hfil \begin{tabular}{@{}c@{}}Recall \\ \scriptsize{(macro, weighted)}\end{tabular} & 
            \hfil \begin{tabular}{@{}c@{}}F1-Score \\ \scriptsize{(macro, weighted)}\end{tabular} \\
            
            \hline
            & \hfil Chinese & \hfil 0.907 & \hfil 0.891, 0.909 & \hfil 0.901, 0.907 & \hfil 0.895, 0.907\\ 
             & \hfil English & \hfil 0.838 & \hfil 0.852, 0.852 & \hfil 0.836, 0.838 & \hfil 0.834, 0.838\\ 
             & \hfil French & \hfil 0.858 & \hfil 0.868, 0.867 & \hfil 0.856, 0.858 & \hfil 0.855, 0.858\\
            \hfil 1000 & \hfil German & \hfil 0.926 & \hfil 0.926, 0.926 & \hfil 0.927, 0.926 & \hfil 0.926, 0.926\\
             & \hfil Italian & \hfil 0.750 & \hfil 0.754, 0.756 & \hfil 0.755, 0.750 & \hfil 0.742, 0.750\\
             & \hfil Japanese & \hfil 0.783 & \hfil 0.789, 0.792 & \hfil 0.783, 0.783 & \hfil 0.780, 0.783\\
             & \hfil Russian & \hfil 0.792 & \hfil 0.806, 0.797 & \hfil 0.793, 0.792 & \hfil 0.797, 0.792\\
             & \hfil Spanish & \hfil 0.851 & \hfil 0.850, 0.853 & \hfil 0.824, 0.851 & \hfil 0.830, 0.851\\
             
             \hline
            & \hfil Chinese & \hfil 0.929 & \hfil 0.929, 0.930 & \hfil 0.908, 0.929 & \hfil 0.917, 0.929\\ 
             & \hfil English & \hfil 0.896 & \hfil 0.903, 0.904 & \hfil 0.895, 0.896 & \hfil 0.895, 0.896\\ 
             & \hfil French & \hfil 0.951 & \hfil 0.951, 0.951 & \hfil 0.951, 0.951 & \hfil 0.951, 0.951\\
            \hfil 5000 & \hfil German & \hfil 0.961 & \hfil 0.961, 0.961 & \hfil 0.961, 0.961 & \hfil 0.961, 0.961\\
             & \hfil Italian & \hfil 0.908 & \hfil 0.911, 0.910 & \hfil 0.908, 0.908 & \hfil 0.909, 0.908\\
             & \hfil Japanese & \hfil 0.900 & \hfil 0.901, 0.902 & \hfil 0.899, 0.900 & \hfil 0.900, 0.900\\
             & \hfil Russian & \hfil 0.880 & \hfil 0.891, 0.886 & \hfil 0.873, 0.880 & \hfil 0.879, 0.880\\
             & \hfil Spanish & \hfil 0.933 & \hfil 0.933, 0.933 & \hfil 0.921, 0.933 & \hfil 0.926, 0.933\\
             
             \hline 
             \hfil 5216 & \hfil Russian & \hfil 0.886 & \hfil 0.893, 0.889 & \hfil 0.881, 0.886 & \hfil 0.886, 0.886\\
             
             \hline 
             \hfil 9458 & \hfil Spanish & \hfil 0.933 & \hfil 0.931, 0.937 & \hfil 0.931, 0.933 & \hfil 0.929, 0.933\\
             
            \hline
            & \hfil Chinese & \hfil 0.937 & \hfil 0.945, 0.939 & \hfil 0.907, 0.937 & \hfil 0.923, 0.937\\ 
             & \hfil English & \hfil 0.970 & \hfil 0.970, 0.970 & \hfil 0.970, 0.970 & \hfil 0.970, 0.970\\ 
             & \hfil French & \hfil 0.962 & \hfil 0.962, 0.962 & \hfil 0.962, 0.962 & \hfil 0.962, 0.962\\
            \hfil 10000 & \hfil German & \hfil 0.968 & \hfil 0.968, 0.968 & \hfil 0.968, 0.968 & \hfil 0.968, 0.968\\
             & \hfil Italian & \hfil 0.905 & \hfil 0.909, 0.909 & \hfil 0.905, 0.905 & \hfil 0.905, 0.905\\
             & \hfil Japanese & \hfil 0.916 & \hfil 0.915, 0.917 & \hfil 0.915, 0.916 & \hfil 0.916, 0.916\\
             
             \hline
        \end{tabular}
    \end{center}
\end{table*}

\end{document}